# PO-GVINS: Tightly Coupled GNSS-Visual-Inertial Integration with Pose-Only Representation

Zhuo Xu, Feng Zhu, Zihang Zhang, Chang Jian, Jiarui Lv, Yuantai Zhang, Xiaohong Zhang

*Abstract*—Accurate and reliable positioning is crucial for perception, decision-making, and other high-level applications in autonomous driving, unmanned aerial vehicles, and intelligent robots. Given the inherent limitations of standalone sensors, integrating heterogeneous sensors with complementary capabilities is one of the most effective approaches to achieving this goal. In this paper, we propose a filtering-based, tightly coupled global navigation satellite system (GNSS)-visual-inertial positioning framework with a pose-only formulation applied to the visual-inertial system (VINS), termed PO-GVINS. Specifically, multiple-view imaging used in current VINS requires a priori of 3D feature, then jointly estimate camera poses and 3D feature position, which inevitably introduces linearization error of the feature as well as facing dimensional explosion. However, the pose-only (PO) formulation, which is demonstrated to be equivalent to the multiple-view imaging and has been applied in visual reconstruction, represent feature depth using two camera poses and thus 3D feature position is removed from state vector avoiding aforementioned difficulties. Inspired by this, we first apply PO formulation in our VINS, i.e., PO-VINS. GNSS raw measurements are then incorporated with integer ambiguity resolved to achieve accurate and drift-free estimation. Extensive experiments demonstrate that the proposed PO-VINS significantly outperforms the multi-state constrained Kalman filter (MSCKF). By incorporating GNSS measurements, PO-GVINS achieves accurate, drift-free state estimation, making it a robust solution for positioning in challenging environments.

*Index Terms*—multi-sensor fusion navigation, pose-only formulation, tightly coupling, global navigation satellite system, visual-inertial system

## I. Introduction

Continuous, reliable, and high-precision positioning is essential for emerging applications, such as unmanned ariel vehicles (UAVs), autonomous cars, and other intelligent robots. However, due to inherent limitations of standalone sensors, multi-sensor fusion, which incorporates complementary information from heterogeneous sensors, is considered an effective means of addressing this issue.

Exploring unknown environments with a monocular camera has received lots of research attention, namely simultaneous localization and mapping (SLAM), due to its cheap, low-cost, and ample semantic information. However, it suffers from scale ambiguity, motion blur, and illumination change, leading to incapable use of estimates for robots. A feasible method is to incorporate inertial measurement unit (IMU), which provides rigorous dynamic model by observing motion of carriers and is independent on external infrastructures, enabling to recover the metric scale [1], [2], as well as to resist the degenerate cases of a single camera. The visual-inertial navigation system (VINS) has also received lots of research attention and been applied in spacecraft landing and descent [3], [4], and unmanned ariel vehicle navigation [5]. Nevertheless, it has been proven that VINS has four unobservable directions, where drifts will be accumulated along the four directions. Complementarily, global navigation satellite system (GNSS) provides absolute and high-precision measurements from navigation satellites, which can theoretically achieve centimeter-level positioning in open-area environments. On one hand, fusing GNSS measurements can easily eliminate drift error caused by VINS and ensures drift-free positioning results. On the other hand, the performance of GNSS rapidly degrades due to its vulnerable features under urban viaducts and other signal blocked areas, where VINS is capable to maintain acceptable positioning accuracy.

Currently, frameworks of fusing GNSS, IMU, and monocular camera can be categorized into the loosely coupled and the tightly coupled. Loosely coupled methods generally regard GNSS and VINS as independent black boxes and directly integrate their estimates. While this approach is time-efficient, it overlooks potential GNSS information, particularly in scenarios where the number of tracked satellites falls below four. This limitation often results in unnecessary accumulation of errors in VINS systems. Conversely, tightly coupled methods make full use of raw measurements from heterogeneous sensors, and thus it enables to achieve more accurate and consistent state estimation [6]. From the perspective of efficiency, in addition to constant dimension of state vector, i.e., camera poses in sliding window, time complexity is severely dependent on the number of correctly matched features in VINS. On one hand, rich textures in surroundings make VINS more robust and can relieve accumulation of errors. On the other hand, dimensional explosion occurs due to additional parameters added to the state vector, i.e., position of each feature point [7]. To this end, researchers employed inverse depth expression in their fusion

This study was supported by the National Key Research and Development Program of China (Grant No.2022YFB3903802), the National Science Fund for Distinguished Young Scholars of China (Grant No. 42425003), the National Natural Science Foundation of China (Grant No. 42388102), the National Natural Science Foundation of China (Grant No. 42374031) and the National Natural Science Foundation of China (Grant No. 42104021).

Zhuo Xu, Feng Zhu, Zihang Zhang, Chang Jian, Jiarui Lv, Yuantai Zhang, Xiaohong Zhang are with the School of Geodesy and Geomatics, Wuhan University, Wuhan, Hubei 430079, China. E-mail: {zhuoxu, fzhu, zihangzhang, 2024202140027, lvjiauri, officialtai}@whu.edu.cn, xhzhang@sgg.whu.edu.cn.

Feng Zhu is also with Hubei Luojia Laboratory, Wuhan University, Wuhan, Hubei 430079, China.

Xiaohong Zhang is also with Chinese Antarctic Center of Surveying and Mapping, Wuhan University, Wuhan, Hubei 430079, China

(Corresponding author: Feng Zhu)

framework, reducing the dimension of each feature point from 3 to 1 [8]. Furthermore, MSCKF projects visual bearing measurements onto the null space of the feature Jacobian matrix, which explicitly eliminates feature points during state estimation [2]. In their implementation, however, an empirical threshold is used to restrict the number of the maximum features during state estimation to avoid potential dimensional explosion. And the null space projection operates *after* linearization, ignoring the higher order terms. Recent research indicated that visual pose-only (PO) formulation expresses feature depth by two camera poses, eliminating parameters of feature points *before* linearization, and demonstrated its equivalence with traditional multi-view geometry, providing a new perspective to balance accuracy and efficiency.

From the view of GNSS observation model, researchers have explored tightly fusion of standard point positioning (SPP)/VINS, precise point positioning (PPP)/VINS. SPP uses pseudorange and Doppler measurements, which only achieves meter-level positioning accuracy. The PPP can potentially achieve centimeter-level accuracy due to the use of phase measurements with ambiguity resolution (AR). However, PPP needs a long convergence time, and AR is not deeply discussed in these integrated frameworks [9]. Real-time kinematics (RTK) has advantages of instant convergence and is easy to achieve continuous AR of carrier phase measurements by using double-differenced GNSS observations. [6] has deeply explored RTK-based fusion approaches with AR, yet their VINS still suffers from dimensional explosion or ignores higher order terms when using null space projection. Although the cutting-edge PPP-RTK model has also explored in GNSS/VINS framework, RTK has a more concise form and thus it is the most widely used GNSS formulation.

To this end, this paper proposes a tightly integrated RTK-visual-inertial framework, where visual measurements are reformulated by PO model, namely PO-GVINS. In brief, the main contribution of this work includes:

a) We propose a filtering-based and tightly coupled framework to integrate raw measurements from GNSS-RTK (both pseudorange and carrier phase), IMU and a monocular camera with continuous integer ambiguity resolved.

b) The PO representation is applied in our visual observation model to avoid dimensional explosion and linearization errors introduced when using null space projection in MSCKF. Thus, the PO formulation mitigates accumulative error, and further ensures an accurate estimation.

c) The performance of proposed PO-GVINS is evaluated through private dataset with challenging GNSS degenerated scenarios. The results show that PO formulation shows significant improvements compared with MSCKF, and the proposed PO-GVINS achieves a more accurate and drift-free estimation.

## II. Related Work

Researches in GNSS/IMU/camera have flourished in the past five years and numerous enlightening methods have been explored. Relevant literature is revisited from VINS and heterogeneous sensors fusion respectively.

### A. Visual-inertial navigation system

Tightly coupled frameworks, where parameters of cameras and IMUs are joint estimated using raw measurements from each sensor, showing better accuracy [10]. A time efficient and profoundly influential framework adopts a sliding window filtering (SWF) to constrain multiple cameras and project visual bearing measurements onto the null space of the feature Jacobian matrix, rather than estimate them directly, namely MSCKF [2], [11]. Based on MSCKF, researchers further utilize mapped landmarks as well as tracked ones in spacecraft landing and descent [12], apply observability constraints to avoid wrong observability property for fast UAVs [13], [14]. Besides, other filtering methods, such as unscented Kalman filter (UKF), particle filter (PF) et al., are employed to handle the high degree of nonlinearity and non-Gaussian noise [15], [16], [17], [18], [19]. In [20], [21], invariant EKF (InEKF) has been introduced to preserve the observable structure of the navigation system. Besides, SchurVINS also utilizes stochastic clone to maintain a sliding window, but it utilizes Schur complement to build equivalent residual equation, and position of feature position can be recovered [22]. Yet the equivalent between Schur complement and null space projection has been demonstrated in [23], SchurVINS still suffers from ignoring higher order terms as stated before.

Another way to integrate IMU and camera measurements is to use graph optimization, where preintegration is utilized to process IMU measurements [24]. The earliest tightly coupled VIO system can be traced back to OKVIS [25]. VINS-mono is then proposed with loop-closure and provided ability to online estimate both intrinsic and extrinsic parameters [1]. Stereo camera is further supported in VINS-fusion [26]. Based on [27] and [28], ORBSLAM 3 is proposed which supports both monocular and stereo camera measurements fused with IMU preintegration measurements. Similarly, SVO originally supports visual and multiple cameras. And currently it has released supports of inertial sensors based on optimization [29]. In addition to indirect and semi-direct methods, [30] applied direct methods with stereo and inertial tightly integrated system. Sparse odometry (DSO)-VIO proposed a dynamic marginalization strategy in order to keep VINS consistent. [31]. And [32] proposes a delayed marginalization methods to inject IMU information into already marginalized states based on direct formulation.

Although there are works focusing on keeping system sparsity when marginalize old states [33], [34], reducing dimension of feature representation [8], and reusing previous calculations [35], both filtering-based and optimization-based approaches still suffer from tradeoff between efficiency and accuracy. Recently, a visual pose-only (PO) representation has been proposed, which explicitly eliminates feature depth in visual measurement equation, and its equality with multiple geometry has been demonstrated [36], [37].

### B. Integration of GNSS, IMU, and visual measurements

In order to eliminate accumulated drifts in VINS, GNSS is initially incorporated as a black box, where position of antennas estimated by GNSS is integrated, i.e. loosely coupled [26], [38], [39], [40]. However, these loosely coupled methods cannot take full use of GNSS raw measurements, especially in those signal blocked scenarios [6]. On the contrary, tightly coupled manner provides the ability to fuse GNSS raw measurements even when the visible satellites are less than 4. [41] and [42] respectively utilize filtering-based and optimization-based framework to fuse raw measurements, but only pseudorange and doppler measurements are utilized, limiting accurate performance of GNSS. Recent researches further utilize high-precision phase measurements and ambiguity resolution techniques, enabling centimeter-level accuracy, where GNSS is respectively modelled as precise point positioning (PPP) [9], [43], [44], real-time kinematics (RTK) [6], [45], [46], and even the cutting-edge PPP-RTK technology [47], [48] with integer ambiguity resolved. However, PPP suffers a long-time convergence [6] and PPP-RTK is currently a developing technology, making RTK still the most well used and matured GNSS approach.

## III. VISUAL-INERTIAL ODOMETRY WITH POSE-ONLY FORMULATION

### A. Pose-only formulation

Assume a feature point has been observed by $n$ images, where $\boldsymbol{P}^w = (x^w, y^w, z^w)^T$ denotes feature position in world frame and $\boldsymbol{p}_i = (x_i, y_i, 1)^T$ denotes the normalized image coordinate of this feature expressed in i-th view ($i = 1, 2, \cdots, n$), satisfying

$$\boldsymbol{p}_i = \frac{1}{z_i^c}\boldsymbol{p}_i^c = \frac{1}{z_i^c}\boldsymbol{R}_i^T\left(\boldsymbol{P}^w - \boldsymbol{r}_i\right), i = 1, 2, \cdots, n \quad (1)$$

where $\boldsymbol{p}_i^c = (x_i^c, y_i^c, z_i^c)^T$ is the coordinate of this feature point in i-th frame and $(\boldsymbol{R}_i, \boldsymbol{r}_i)$ is the corresponding transformation from i-th camera frame to world frame.

Regarding two base frames $i$ and $j$, the traditional two-view geometry can be expressed as

$$z_j^c \boldsymbol{p}_j = z_i^c \boldsymbol{R}_i^j \boldsymbol{p}_i + \boldsymbol{t}_{j,i} \quad (2)$$

where $\boldsymbol{R}_i^j = \boldsymbol{R}_j^T \boldsymbol{R}_i$, $\boldsymbol{t}_{j,i} = \boldsymbol{R}_j^T(\boldsymbol{r}_i - \boldsymbol{r}_j)$ denotes the local transformation.

To derive pose-only formulation, left multiply the antisymmetric matrix $\boldsymbol{p}_j^\wedge$ on the both side of (2):

$$-\boldsymbol{p}_j^\wedge \boldsymbol{t}_{j,i} = z_i^c \boldsymbol{p}_j^\wedge \boldsymbol{R}_i^j \boldsymbol{p}_i \quad (3)$$

Take the magnitude, then depth $z_i^c$ can be expressed by a pair of frame poses:

$$z_i^c = \frac{\|\boldsymbol{p}_j^\wedge \boldsymbol{t}_{j,i}\|}{\theta_{i,j}} \triangleq d_i^{(i,j)}, \theta_{i,j} \triangleq \|\boldsymbol{p}_j^\wedge \boldsymbol{R}_i^j \boldsymbol{p}_i\| \quad (4)$$

Similarly, left multiplying $(\boldsymbol{R}_i^j \boldsymbol{p}_i)^\wedge$ on both side of (2), and then taking the magnitude, $z_j^c$ can be also expressed by the same pair of poses:

$$z_j^c = \frac{\|(\boldsymbol{R}_i^j \boldsymbol{p}_i)^\wedge \boldsymbol{t}_{j,i}\|}{\|(\boldsymbol{R}_i^j \boldsymbol{p}_i)^\wedge \boldsymbol{p}_j\|} = \frac{\|(\boldsymbol{R}_i^j \boldsymbol{p}_i)^\wedge \boldsymbol{t}_{j,i}\|}{\theta_{i,j}^k} \triangleq d_j^{(i,j)} \quad (5)$$

Thus, (2) can be rewritten as:

$$d_j^{(i,j)} \boldsymbol{p}_j = d_i^{(i,j)} \boldsymbol{R}_i^j \boldsymbol{p}_i + \boldsymbol{t}_{j,i} \quad (6)$$

For any other l-th frame ($l \neq j$), we also have:

$$d_l^{(i,l)} \boldsymbol{p}_l = d_i^{(i,l)} \boldsymbol{R}_i^l \boldsymbol{p}_i + \boldsymbol{t}_{l,i} \quad (7)$$

According to definition, $d_i^{(i,l)} = d_i^{(i,j)} = z_i^c$, we substitute $d_i^{(i,l)}$ with $d_i^{(i,j)}$, deriving PO formulation of the feature point:

$$d_l^{(i,l)} \boldsymbol{p}_l = d_i^{(i,j)} \boldsymbol{R}_i^l \boldsymbol{p}_i + \boldsymbol{t}_{l,i} \quad (8)$$

Comparing with MSCKF's formulation, for example, the linearized formulation of (1) is:

$$\boldsymbol{e}^{(i)} = \boldsymbol{H}_x^{(i)} \delta \boldsymbol{x}_p^{(i)} + \boldsymbol{H}_f^{(i)} \delta \boldsymbol{x}_f^{(i)} + \boldsymbol{n}^{(i)} \quad (9)$$

where, $\boldsymbol{H}_x$ and $\boldsymbol{H}_f$ are Jacobis of camera poses and feature point respectively, and $\delta \boldsymbol{x}_p$, $\delta \boldsymbol{x}_f$ denote error state of camera poses and position of feature depth. $\boldsymbol{r}$ defines the residual and $\boldsymbol{n}$ denotes Gaussian noise. Then, by projecting $\boldsymbol{e}^{(i)}$ on the left null space of the matrix $\boldsymbol{H}_f^{(i)}$, i.e., MSCKF, $\delta \boldsymbol{x}_f^{(i)}$ is also eliminated, and thus a new residual can be derived [2]:

$$\boldsymbol{e}_o^{(i)} = \boldsymbol{H}_{ox}^{(i)} \delta \boldsymbol{x}_p^{(i)} + \boldsymbol{n}_o^{(i)} \quad (10)$$

It is important to highlight that, this null space projection only eliminates first order terms of the depth of feature point. Higher order terms are ignored during linearization. The pose-only formulation, however, is lossless when eliminating feature depth. Besides, camera poses are updated during iteration, thus the feature depth is also implicitly updated. While, it is difficult to recover feature depth in MSCKF, and even if the iteration strategy is applied, linearization point of 3D feature position maintains the original.

### B. Propagation and augmentation

IMU measurements are utilized to integrate and propagate, which is modelled as mechanization. The motion model can be described by

$$\begin{cases} \delta \dot{\boldsymbol{r}}_b = \delta \boldsymbol{v}_b + \boldsymbol{n}_r \\ \delta \dot{\boldsymbol{v}}_b = \boldsymbol{N} \delta \boldsymbol{r}_b - 2(\boldsymbol{\omega}_{ie}^w)^\wedge \delta \boldsymbol{v}_b + (\boldsymbol{f}^w)^\wedge \boldsymbol{\phi} + \boldsymbol{R}_b \delta \boldsymbol{b}_a + \boldsymbol{n}_v \\ \dot{\boldsymbol{\phi}} = -(\boldsymbol{\omega}_{ie}^w)^\wedge \boldsymbol{\phi} - \boldsymbol{R}_b \delta \boldsymbol{b}_g + \boldsymbol{n}_\phi \\ \dot{\boldsymbol{b}}_a = \boldsymbol{\eta}_a, \dot{\boldsymbol{b}}_g = \boldsymbol{\eta}_g \end{cases} \quad (11)$$

Where, $\delta \boldsymbol{x}$ denotes error state of a variable $\boldsymbol{x}$, which is defined by $\delta \boldsymbol{x} = \tilde{\boldsymbol{x}} - \boldsymbol{x}$ and $\tilde{\boldsymbol{x}}$ denotes an observation of true state $\boldsymbol{x}$. Specifically, $\boldsymbol{r}_b$, $\boldsymbol{v}_b$ denote position and velocity of body frame with respect to world frame. Subscript $i$ stands for Earth-centered inertial (ECI) frame. $\boldsymbol{f}^w = \boldsymbol{R}_b \boldsymbol{f}^b$ denotes the accelerometers measurements. And $\boldsymbol{\omega}$ denotes angular velocity. $\boldsymbol{N}$ is the tensor of the gravitational gradients. Besides, $\boldsymbol{n}$ denotes the white Gaussian noise. Accelerometer and gyroscope biases, $\boldsymbol{b}_a$ and $\boldsymbol{b}_g$, are modelled as random walk process, $\boldsymbol{\eta}_a$ and $\boldsymbol{\eta}_g$ respectively. Additionally, error state of the rotation matrix can be obtained by $\boldsymbol{R}_b = \left(\boldsymbol{I} + \boldsymbol{\phi}^\wedge\right)\tilde{\boldsymbol{R}}_b$. In our implementation, world frame is chosen as Earth frame, and thus, $\boldsymbol{\omega}_{ie}^w = \boldsymbol{\omega}_{ie}^e$ is a constant denoting the rotation rate of the Earth. The initialization in e-frame of the whole system will be introduced in section IV.

Then, (11) can be rewritten as matrix form

$$\delta \dot{\boldsymbol{x}}_{imu} = \boldsymbol{F}_t \delta \boldsymbol{x}_{imu} + \boldsymbol{G}_t \boldsymbol{w}. \quad (12)$$

where, IMU error state vector $\delta x_{imu}$ and noise vector $w$ have the following form

$$\delta x_{imu} = \begin{bmatrix} \delta r_b^T & \delta v_b^T & \phi^T & \delta b_a^T & \delta b_g^T \end{bmatrix}^T,$$
$$w = \begin{bmatrix} n_r^T & n_v^T & n_\phi^T & n_a^T & n_g^T \end{bmatrix}^T \quad (13)$$

When camera captures a new image at timestamp $t$, system error state is extended by stochastic clone [49]

$$\delta X_t = \begin{bmatrix} \delta x_{imu} \\ \delta x_{t,imu} \end{bmatrix} = \begin{bmatrix} I_{15\times 15} \\ F_{imu} \end{bmatrix} \delta x_{imu} \quad (14)$$

where, $\delta x_{t,imu} = (\delta r_{t,b}^T, \phi_t^T)^T$ denotes the clone of IMU error state at $t$. $F_{imu}$ is thus defined by

$$F_{imu} = \begin{bmatrix} I_{3\times 3} & 0_{3\times 3} & 0_{3\times 3} & 0_{3\times 6} \\ 0_{3\times 3} & 0_{3\times 3} & I_{3\times 3} & 0_{3\times 6} \end{bmatrix} \quad (15)$$

Then, prior constraints can be derived by covariance propagation law

$$D_{t|t-1} = \begin{bmatrix} D_{t-1} & \left(F_{imu} D_{t-1}\right)^T \\ F_{imu} D_{t-1} & F_{imu} D_{t-1} F_{imu}^T \end{bmatrix} \quad (16)$$

where, $D_{t-1}$ denotes the covariance of the previous propagated system states. Thus, the state vector after augmentation is defined as

$$X_{vins} = \begin{bmatrix} \delta x_{imu}^T, \delta x_{1,imu}^T, \cdots, \delta x_{t,imu}^T, \cdots, \delta x_{N,imu}^T \end{bmatrix}^T \quad (17)$$

where $N$ denotes the window size.

### C. Measurement update

The PO residual $e_l^{po}$ is defined as

$$e_l^{po} = K \frac{d_i^{(i,j)} R_i^l p_i + t_{l,i}}{e_3^T \left(d_i^{(i,j)} R_i^l p_i + t_{l,i}\right)} - \tilde{u}_l \quad (18)$$

where $e_3^T = (0,0,1)$, $K$, $\tilde{u}_l$ define the camera intrinsic parameters and monocular on image plane respectively. (18) can be further simplified considering the definition of $d_i^{(i,j)}$:

$$e_l^{po} = K \frac{\left\| p_j^\wedge t_{j,i} \right\| R_i^l p_i + \theta_{i,j} t_{l,i}}{e_3^T \left(\left\| p_j^\wedge t_{j,i} \right\| R_i^l p_i + \theta_{i,j} t_{l,i}\right)} - \tilde{u}_l \triangleq K \frac{Y_l}{e_3^T Y_l} - \tilde{u}_l \quad (19)$$

Note that our state vector contains IMU poses instead of the camera ones, the transformation from c-frame to b-frame can be obtained by extrinsic parameters:

$$r_b = r_c - R_b^e l_c, \quad R_b^e = R_c^e R_b^c \quad (20)$$

Where $l_c$ denotes the camera position with respect to b-frame, and $R_b^c$ rotates from b-frame to c-frame. Jacobian matrix can be derived according to the chain rule (21).

Thereafter, these measurements can be used to EKF update with constrained by $D_{t|t-1}$.

As aforementioned, $(i,j)$ is a pair of base frames. Considering $\theta_{\eta,\xi}$ which indicates a quality indicator [37], $\theta_{\eta,\xi}$ ($1 \leq \eta, \xi \leq n, \eta \neq \xi$) is calculated for all candidates, and the maximum yields our base frames.

$$J_l^{(i)} = \left(\frac{\partial e_l^{po}}{\partial \delta r_i} \quad \frac{\partial e_l^{po}}{\partial \phi_i}\right) = \frac{\partial e_l^{po}}{\partial Y_l}\left(\frac{\partial Y_l}{\partial \delta r_i} \quad \frac{\partial Y_l}{\partial \phi_i}\right),$$

$$J_l^{(j)} = \left(\frac{\partial e_l^{po}}{\partial \delta r_j} \quad \frac{\partial e_l^{po}}{\partial \phi_j}\right) = \frac{\partial e_l^{po}}{\partial Y_l}\left(\frac{\partial Y_l}{\partial \delta r_j} \quad \frac{\partial Y_l}{\partial \phi_j}\right), \quad (21)$$

$$J_l^{(l)} = \left(\frac{\partial e_l^{po}}{\partial \delta r_l} \quad \frac{\partial e_l^{po}}{\partial \phi_l}\right) = \frac{\partial e_l^{po}}{\partial Y_l}\left(\frac{\partial Y_l}{\partial \delta r_l} \quad \frac{\partial Y_l}{\partial \phi_l}\right)$$

## IV. TIGHTLY INTEGRATION OF GNSS, VISUAL AND INERTIAL RAW MEASUREMENTS

In order to eliminate accumulated errors introduced by VINS, measurements from GNSS, including both pseudorange and phase observations, are sequentially incorporated and then double-differenced integer ambiguity is resolved using LAMBDA, enabling high-precision positioning.

Considering signals from satellite $s$, its pseudorange and phase observations can be modelled as

$$\begin{cases} L_r^s = \rho_r^s + cdt_r - cdt^s + I_r^s + T_r^s + n_r^s \\ \phi_r^s = \rho_r^s + cdt_r - cdt^s + \lambda \cdot N_r^s - I_r^s + T_r^s + \varepsilon_r^s \end{cases} \quad (22)$$

Where, $\phi_r^s$, $L_r^s$ are pseudorange and phase measurements respectively. $dt_r$, $dt^s$ denote clock error of receiver and satellite. $I_r^s$, $T_r^s$ denote ionospheric and tropospheric delay. $\lambda$ denotes the wavelength and $N_r^i$ denotes the carrier phase ambiguity. $e_r^s$, $\varepsilon_r^s$ denote white Gaussian noise of pseudorange and phase observations. $\rho_r^s$ defines the distance between receiver and satellite, derived by (23), where $l_a$ denotes the vector pointing from IMU center to GNSS antenna and $\hat{r}_b$, $\hat{R}_b$ denotes current estimates of IMU pose derived by VINS.

$$\rho_r^s = \left\| \hat{r}_b - r_s + \hat{R}_b l_a \right\|_2 \quad (23)$$

Regarding a stationary station $u$ observing the same satellite $s$, the single-differenced observations can be derived by (24), which eliminates satellite clock error.

$$\begin{cases} \Delta L_{ru}^s = L_r^s - L_u^s \\ \quad = \rho_{ru}^s + dt_{ru} + I_{ru}^s + T_{ru}^s + n_{ru}^s \\ \Delta \phi_{ru}^s = \phi_{ru}^s - \phi_{ru}^s \\ \quad = \rho_{ru}^s + dt_{ru} + \lambda_f \cdot N_{ru}^s - I_{ru}^s + T_{ru}^s + \varepsilon_{ru}^s \end{cases} \quad (24)$$

Select a reference satellite $k$ and then the double-differenced (DD) observations can be further derived by (25), which is also known as RTK formulation (short baseline scenario).

$$\begin{cases} \nabla \Delta L_{ru}^{sk} = \Delta L_{ru}^s - \Delta L_{ru}^k = \rho_{ru}^{sk} + n_{ru,f}^{sk} \\ \nabla \Delta \phi_{ru}^{sk} = \Delta \phi_{ru}^s - \Delta \phi_{ru}^k = \rho_{ru}^{sk} + \lambda \cdot N_{ru}^{sk} + \varepsilon_{ru,f}^{sk} \end{cases} \quad (25)$$

In addition to VINS state $X_{vins}$, DD ambiguities are directly appended to the state vector and the priori can be derived by

$$X_{gvins} = \begin{bmatrix} X_{vins}^T, N_{ru} \end{bmatrix}^T$$
$$D_{gvins} = \begin{bmatrix} D_t & 0 \\ 0 & D_{amb} \end{bmatrix} \quad (26)$$

where $N_{ru}$ defines a vector containing all DD ambiguities at current timestamp, and $D_{amb}$ denotes the prior covariance of DD ambiguities, which can be propagated from undifferenced ambiguities. $D_t$ denotes the posteriori from VINS.

Then, GNSS residual and the Jacobi can be easily derived from (25), and then an iterative EKF update is processed with IGG-III outlier removal strategy applied. After update, DD ambiguities are resolved through LAMBDA method.

Besides, in our implementation, a robust cascaded alignment is first conducted to initialize GNSS/IMU system [50], since GNSS provides absolute and high-precision positioning ability. Then, initial IMU pose, $r_b$ and $R_b$, can be determined. With known extrinsic parameters, camera poses can also be recovered with metric scale.

## V. Experimental Results

For comparison, we implement Pose-Only VINS (PO-VINS), MSCKF, GNSS/IMU (GI), MSCKF (M)-GVINS, and the proposed PO-GVINS with our own platform. When process the following experiments, the same strategies are utilized, e.g., in our implementation of both PO-VINS and MSCKF, the same visual feature selection approach and the maximum allowed number of iterations is kept same to truly reflect the performance of different observation model.

*1) Experimental Setup:* As shown in Fig. 1, we evaluate the proposed PO-GVINS using data collected by a self-developed hardware, which includes a GNSS receiver together with its antenna, four monocular cameras (front-view camera is used in this research), and a MESE IMU. Besides, a high-precision IMU (ISA-100C) is utilized to obtain reference trajectory. All these sensors are hard synchronized and extrinsic parameters relative to MEMS IMU are well calibrated. The base station, required by RTK algorithm, is equipped with a Trimble Alloy receiver, and its antenna is set at an open-sky view. The distance between rover and base station is no greater than 10km to meet the requirement of short baseline. In our test, reference trajectory is solved through the smoothed and combined solutions of multi-GNSS RTK/IMU(ISA-100C), using commercial software Inertial Explorer (IE) 8.9. MESE-IMU is used to evaluate our proposed PO-GVINS. The specifications of each important sensor are listed in TABLE I.

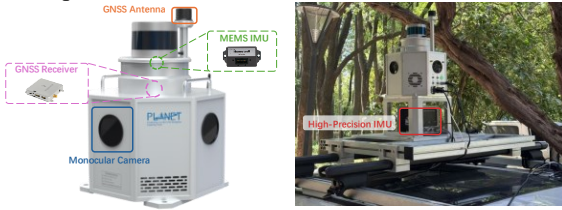

Fig. 1. Experimental hardware (left) and collection vehicle (right).

We then conduct a large-scale experiment, where the length of the trajectory is over 10km, covering both visual-challenging and GNSS-challenging scenarios, as shown in Fig. 2. illustrates the error sequences of GNSS-RTK/IMU (GI) in Right (R), Front (F) and Up (U), and the shaded regions correspond to test A and B. Fig. 3 also shows the visible satellites and position dilution of precision (PDOP), which are two indicators describing GNSS quality. A smaller number of satellites and a larger PDOP value generally indicate poor GNSS observation quality.

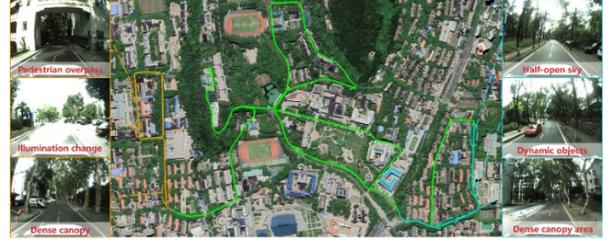

Fig. 2. Overview of Test A and Test B. (a) Trajectories (green) and typical challenging scenarios in Test A (cyan) and B (orange).

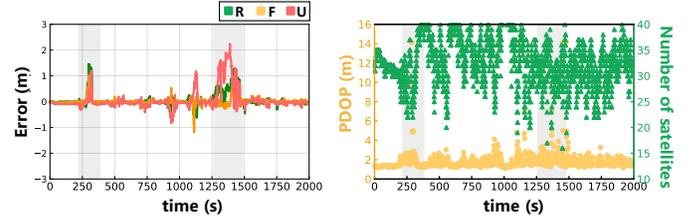

Fig. 3. GNSS/MEMS-IMU tightly coupled solution and GNSS indicators for the whole trajectory. Shaded regions are Test A and B respectively from the left to right.

TABLE I
SENSOR SPECIFICATIONS FOR THE DEVICE USED IN VEHICLE-BORNE EXPERIMENTS

| Sensor Type & Item | Specification |
| --- | --- |
| **High-precision IMU** | **Novatel SPAN-ISA-100C** |
| Gyroscope bias | 0.05 °/$hr$ |
| Gyroscope random walk | 0.005 °/$\sqrt{hr}$ |
| Accelerometer bias | 0.1 mg |
| Measurement frequency | 200 Hz |
| **MEMS-IMU** | **HGuide I300** |
| Gyroscope bias | 65 °/$hr$ |
| Gyroscope random walk | 0.15 °/$\sqrt{hr}$ |
| Accelerometer bias | 1.0 mg |
| Measurement frequency | 200 Hz |
| **Camera** | **MER-131** |
| Shutter | Global shutter |
| Resolution | 1280×1024 pixel |
| Measurement frequency | 10 Hz |
| **GNSS receiver** | **Septentrio Mosaic-X5** |
| Antenna | Harxon HX-CH7609A |
| Measurement frequency | 1 Hz |

*2) Test A:* Fig. 4 intuitively show that the PO-VINS is smoother compared with MSCKF. MSCKF, however, occurs failure at the end of the test. Both VINS approaches suffer from accumulated errors over travelled distances, and thus trajectories gradually diverge from the reference. With the absolute positioning feature of GNSS, PO-GVINS is closer to the reference compared with GI, indicating better estimating accuracy. TABLE II shows the statistics of test A. Compared with MSCKF, the PO representation shows obvious improvements, especially on translation. Since we use the same feature selection methods, feature point linearization error introduced by MSCKF leads to significant increasing errors. The PO formulation, however, eliminates depth of the feature point before linearization. During estimation, camera poses are updated and the feature depth, represented by a pair of camera poses, is also updated without linearization error. Besides, Although GI can achieve drift-free estimation, but it is vulnerable in GNSS challenging scenarios. While, PO-GVINS, incorporating visual measurements, improves 54.6% and 5.3% on translation and rotation RMS compared with M-GVINS.

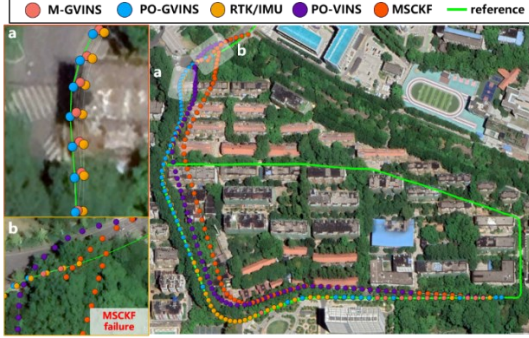

Fig. 4. Estimated trajectories compared with the reference in test A.

Fig. 5 further digs into details of estimating accuracy. Although the number of visible satellites is sufficient, GNSS is affected by surrounding buildings and trees, which leads to multi-path and non-line-of-sight errors (NLOS). This will lead to the increasement of PDOP value. And thus, the positioning error of GI increases when entering the dense canopy area. The proposed PO-GVINS, however, still maintain smooth and high-precision estimation. The cumulative distribution, as shown in Fig. 6, also demonstrates the best performance of PO-GVINS, compared with GI and M-GVINS, and the effectiveness of PO formulation, compared with MSCKF. Specifically, 95% error of PO-GVINS, M-GVINS, GI, PO-VINS, and MSCKF are 0.72m, 1.56, 2.02m, 32.82m, and 36.10m respectively.

TABLE II
STATISTICS IN TEST A

| Methods | Absolute Translation Error[1] (%) | | | Absolute Rotation Error[1] (deg/m) | | |
|---|---|---|---|---|---|---|
| | max | Avg. | RMS | max | Avg. | RMS |
| MSCKF | 9.63 | 4.50 | 5.35 | 0.24 | **0.18** | **0.18** |
| PO-VINS | **4.43** | **2.80** | **3.12** | 0.24 | 0.19 | 0.19 |

| Methods | Absolute Translation Error (m) | | | Absolute Rotation Error (deg) | | |
|---|---|---|---|---|---|---|
| | max | Avg. | RMS | max | Avg. | RMS |
| GI | 2.31 | 0.76 | 1.05 | 1.86 | 1.67 | 1.67 |
| M-GVINS | 1.97 | 0.75 | 0.97 | 1.86 | 1.52 | 1.52 |
| PO-GVINS | **0.86** | **0.38** | **0.44** | 1.86 | **1.44** | **1.44** |

[1] Divided by length of the test trajectory, 753.64m.

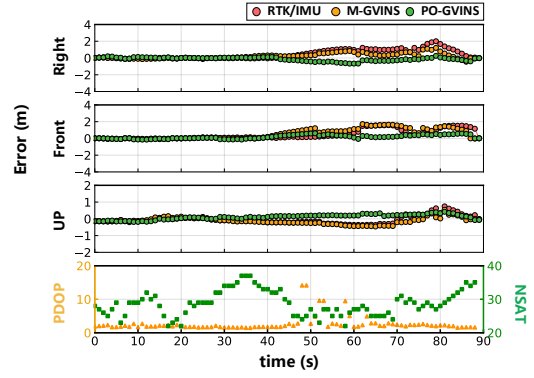

Fig. 5. Error time series and GNSS indicators of test A.

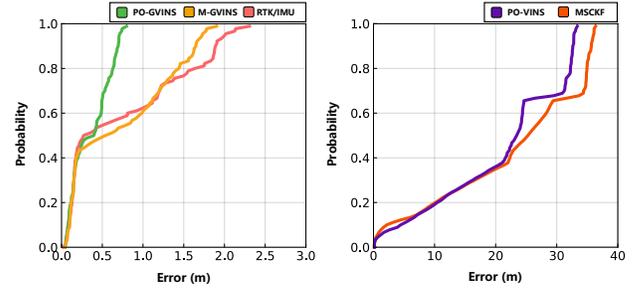

Fig. 6. Cumulative distribution of positioning errors of each configuration in test A.

3) Test B: Test B contains more challenging scenarios including pedestrian overpass and illumination change, which severely affect both GNSS and visual measurements, as shown in Fig. 1. Similarly, Fig. 7 first shows the estimated trajectories, and it is intuitively showing that PO-VINS outperforms MSCKF even at the start of the trajectory. However, due to the accumulative error, both trajectories are biased from the reference. The proposed PO-GVINS, on the contrary, estimated the most consistent trajectory, showing significant improvements compared with PO-VINS and M-GVINS.

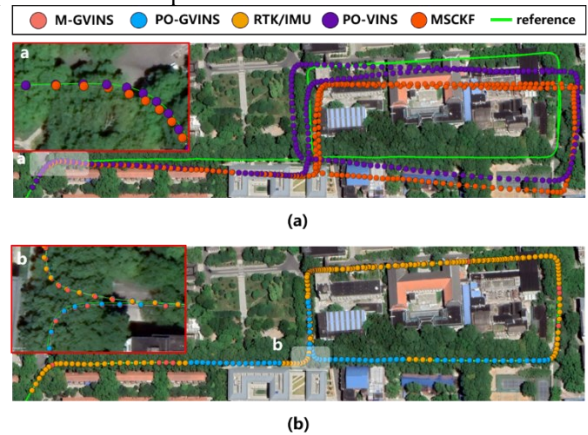

Fig. 7. Estimated trajectories compared with the reference in test B. (a) Comparison between PO-VINS and MSCKF. (b) Comparison among PO-GVINS, M-GVINS, and GI.

TABLE III shows the statistics of test B, which also demonstrates the high-precision positioning ability. The PO-

VINS and PO-GVINS improve 29.8% and 32.3% on absolute translation RMS compared with MSCKF and M-GVINS respectively.

TABLE III
STATISTICS IN TEST B

| Methods | Absolute Translation Error[1] (%) | | | Absolute Rotation Error[1] (deg / m) | | |
|---|---|---|---|---|---|---|
| | max | Avg. | RMS | max | Avg. | RMS |
| MSCKF | 4.51 | 2.06 | 2.25 | 0.11 | 0.09 | 0.09 |
| PO-VINS | **2.27** | **1.48** | **1.58** | **0.11** | **0.09** | **0.09** |
| Methods | Absolute Translation Error (m) | | | Absolute Rotation Error (deg) | | |
| | max | Avg. | RMS | max | Avg. | RMS |
| GI | 2.70 | 0.56 | 0.71 | 1.75 | 1.52 | 1.52 |
| M-GVINS | 2.01 | 0.48 | 0.62 | 1.74 | 1.52 | 1.52 |
| PO-GVINS | **1.45** | **0.36** | **0.42** | **1.58** | **1.46** | **1.46** |

[1] Divided by length of the test trajectory, 1488.68m.

As shown in Fig. 8, the error sequence of GI exists obvious gaps, which is affected by outliers. With visual measurements, the whole system can maintain accurate positioning in those GNSS-challenging scenarios, and this accurate position is helpful to detect GNSS outliers. Furthermore, the proposed PO-GVINS shows a smoother and a more accurate estimated result compared with M-GVINS. Fig. 9 further show the cumulative distribution, the 95% error of PO-GVINS, M-GVINS, GI, PO-VINS, and MSCKF are 0.75m, 1.19m, 1.47m, 33.11m, and 53.59m respectively, showing the best performance of PO-GVINS and effectiveness of PO representation.

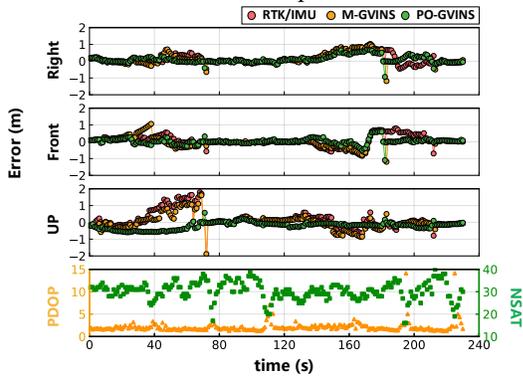

Fig. 8. Error time series and GNSS indicators of test B.

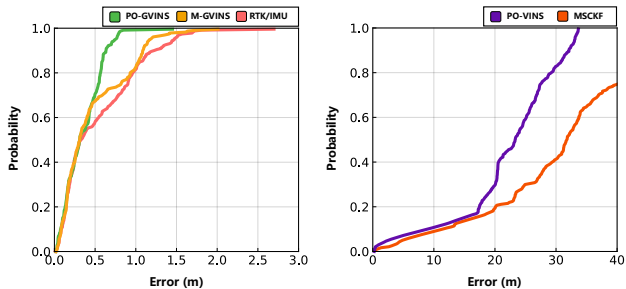

Fig. 9. Cumulative distribution of positioning errors of each configuration in test B.

## VI. CONCLUSION

In this paper, a filtering-based GNSS, IMU, and monocular camera tightly coupled framework is proposed, achieving accurate and drift-free estimation. Specifically, the pose-only formulation is utilized in our visual measurements processing, which avoids feature linearization errors compared with MSCKF. Extensive experiments demonstrate the PO formulation outperforms MSCKF. With the help of visual measurements, the proposed PO-GVINS achieves a more accurate and smoother estimation.